\documentclass[journal]{IEEEtran}
\usepackage{cite}
\usepackage{hyperref}
\usepackage{multirow}

\usepackage{graphicx}
\usepackage{caption}
\ifCLASSINFOpdf
\else
\fi
\usepackage{amsmath}
\usepackage{algorithmic}
\usepackage{algorithm}
\usepackage{array}
\begin{document}
\title{Convolution Neural Network Architecture Learning\\ for Remote Sensing Scene Classification}
\author{
        Jie Chen, Haozhe Huang, Jian Peng, Jiawei Zhu, Li Chen, Wenbo Li, Binyu Sun, Haifeng Li*~\IEEEmembership{Member,~IEEE} % <-this % stops a space
%\thanks{M. Shell was with the Department
%of Electrical and Computer Engineering, Georgia Institute of Technology, Atlanta,
\thanks{Jie Chen, Haozhe Huang, Jian Peng, Jiawei Zhu, Li Chen are with School of Geosciences and Info-Physics, Central South University.}% <-this % stops a space
\thanks{Wenbo Li and Binyu Sun are with Institute of Technology Innovation, Hefei Institutes of Physical Science, Chinese Academy of Sciences.}
\thanks{This work was supported by the National Natural Science Foundation of China (grant numbers 41871364, 41871302, 61773360, and 41671357). Corresponding author: Haifeng Li, Email: lihaifeng@csu.edu.cn}
}
\maketitle

\begin{abstract}
Remote sensing image scene classification is a fundamental but challenging task in understanding remote sensing images. Recently, deep learning-based methods, especially convolutional neural network-based (CNN-based) methods have shown enormous potential to understand remote sensing images. CNN-based methods meet with success by utilizing features learned from data rather than features designed manually. The feature-learning procedure of CNN largely depends on the architecture of CNN. However, most of the architectures of CNN used for remote sensing scene classification are still designed by hand which demands a considerable amount of architecture engineering skills and domain knowledge, and it may not play CNN's maximum potential on a special dataset. In this paper, we proposed an automatically architecture learning procedure for remote sensing scene classification. We designed a parameters space in which every set of parameters represents a certain architecture of CNN (i.e., some parameters represent the type of operators used in the architecture such as convolution, pooling, no connection or identity, and the others represent the way how these operators connect). To discover the optimal set of parameters for a given dataset, we introduced a learning strategy which can allow efficient search in the architecture space by means of gradient descent. An architecture generator finally maps the set of parameters into the CNN used in our experiments.
%Moreover, in order to catch the context information of remote sensing images and obtain better performances, atrous convolution which merely been used in scene classification tasks is also added into the parameter space. 
Experiment results on three remote sensing scene classification benchmarks (UC Merced Land-Use, AID, and NWPU-RESISC45) indicate that the performances of architectures learned from data outperform several classical hand-designed architectures. Different architecture will be good at learning different representation, our proposed architecture learning method potentially help us understand which types of representation are crucial for remote sensing image intelligent understanding.
\end{abstract}


% \fi
\IEEEpeerreviewmaketitle

\section{Introduction}
\IEEEPARstart{W}{ith} the advancement of earth observation techniques, the resolution of remote sensing images (e.g., hyperspectral images \cite{Plaza2011Parallel}, SAR images \cite{Cantalloube2013Airborne}, etc.) has been greatly improved. This allows a more intelligent understanding of remote sensing images and drives a large number of applications, such as classification of land use and land cover (LULC), remote sensing image retrieval, geographic object detection, traffic planning, urban growth measurement, and hazards monitoring  \cite{cheng2017remote,cheng2014multi,wang2016three,zhu2016bag,estoque2015pixel}. These applications mostly demand an intelligent scene classification of remote sensing images, namely giving each scene image a specific label (e.g. freeway, river, grassland) \cite{cheng2017remote}. Traditional pixel-level methods\cite{blaschke2001s} can hardly satisfied the need of remote sensing images scene classification. Object-based image analysis (OBIA) \cite{blaschke2010object} and geographic-object-based image analysis (GEOBIA) \cite{blaschke2014geographic} partition scene image into several segments, but carrying little semantic information which is essential to describe the scenes. For the complexity of remote sensing scenes, how to extract good semantic information for scene classification is a challenging task.

It is well known that good feature descriptions or representations are critical to obtain semantic information from remote sensing images. In other words, feature engineering is one of the most important procedures in remote sensing scene classification. Handcrafted features(such as
%color histograms \cite{Swain1991Color}, texture descriptors \cite{Haralick1973Textural},
GIST \cite{Oliva2001Modeling}, SIFT \cite{Lowe2004Distinctive}, HOG \cite{Dalal2005Histograms}) are earliest used for solving remote sensing scene classification tasks. It is found that designing well-performed handcrafted features is time-consuming and domain knowledge intensive. And there are various scenes in remote sensing images with different spatial resolutions and imaging conditions. Handcrafted features will restrict the capacity of feature representations for that they are universal and may not best fit certain remote sensing images. Recently, deep learning-based\cite{lecun2015deep} methods especially CNN-based methods lead a paradigm shift from designing features manually to learning features from data automatically. And CNN-based methods have achieved better performance than traditional methods in most of the remote sensing images intelligent understanding tasks\cite{scott2017training,chaib2017deep,li2017integrating,paoletti2018new}. CNN learns optimal task-special features from remote sensing images. These features fit data better, and provide high-level information of images which necessary for robust feature representation.

There are two main parts in the feature-learning procedure of CNN-based methods, one is designing the architecture of CNN and the other is training CNN with images. Nogueira \textit{et al.}\cite{nogueira2017towards} found that AlexNet\cite{krizhevsky2012imagenet} performs better than VGG16\cite{simonyan2014very} on the UCMerced Land-Use dataset\cite{yang2010bag} but the result is opposite on the Brazilian Coffee Scenes Dataset\cite{penatti2015deep}. These CNNs with different architectures are trained in the same condition, which indicates the performances are mostly depended on the architectures of CNNs. Note that, previous work often design new architecture of CNN to satisfy the need of remote sensing images intelligent understanding.
However, the architectures designed manually are also enslaved to the domain knowledge of network designers. To design better architectures for a special task or data set, a paradigm shift from designing architectures to learning architectures for data is imperative. In the domain of understanding natural images, 
An attempt in remote sensing is that Chen et al. \cite{chen2019automatic} proposed an automatic CNN design approach in remote sensing HSI image classification.

In this paper, we explore a paradigm shift from designing architectures manually to learning architectures automatically. We also introduced and investigated CNN architectures learned from data for remote sensing scene classification tasks. The major contributions of this article are as follows:

1)	We proposed an architecture-learning procedure to design CNN for remote sensing scene classification tasks.

2)	We add atrous convolution structure into the architecture space to catch a larger context for high-resolution aerial or satellite remote sensing images. Experiments show that all the best architecture learned by our framework contains atrous convolution.

3)	Our experiments show that the best architectures outperform several classical hand-designing architectures in three remote sensing scene classification data sets. Architecture determines the function of CNN, this architecture-learned method may help to understand which representations are important for remote sensing scene classification tasks.

\section{Related Works}
In this section, the related works of remote sensing scene classification and neural architecture search methods are briefly reviewed.
\subsection{Scene Classification}
The previous works about remote sensing scene classification are in varied forms but can be roughly divided into the following three categories according to the features they used:
handcrafted features, unsupervised learning features, and deep learning features \cite{cheng2017remote}.
The methods based on handcrafted features are the earliest in scene classification of a
remote sensing image. Color histograms and texture descriptors \cite{bhagavathy2006modeling}, \cite{dos2010evaluating} are global features that can be sent to the classifier directly. Scale-invariant feature transform (SIFT) \cite{yang2012geographic}, \cite{risojevic2012fusion} and histogram of oriented gradients \cite{cheng2014multi}, \cite{cheng2015auto} are
local features that usually need mid-level descriptor to generate
the entire representation \cite{yang2012geographic}, \cite{zhu2016bag}. Recently, a combination
of multiple different features is considered as a promising
approach to seek further promotion \cite{risojevic2012fusion,zhu2016bag,zou2016scene}. For
example, Zhu et al.\cite{zhu2016bag} propose a local–global feature for
bag-of-visual-words scene classifier, which can combine several features by a fusion operation at the histogram level.
Nevertheless, how to design an effective model to combine
these features is very difficult, and the representation ability
of handcrafted features becomes weaker with the increasing
challenge of this task.

\subsection{Architecture Learning Mechanism}
To design better architectures for a special task or data set, a paradigm shift from designing architectures to learning architectures for data is imperative. In the domain of understanding natural images, 
a great number of different learning strategies, including random search, Bayesian optimization, reinforcement learning (RL), evolutionary methods, and gradient-based methods, have been used to learn the best neural architecture. Among those learning strategies, RL and evolutionary algorithms have attracted most attention and achieved remarkable progress. Zoph and Le \cite{zoph2016neural} proposed an RL-based learning strategy to learning the whole CNN architecture. In the RL-based method, an RNN is built as a controller to generate architectural hyperparameters of neural networks. A CNN with the generated architecture is developed and evaluated. The performance on the validation set is used as the reward to optimize the RNN controller\cite{zoph2018learning}. Evolutionary-based algorithms have also been used for optimizing the neural architecture and achieved comparable results compared with RL-based methods \cite{real2017large,real2019regularized}. In evolutionary-based learning methods, each neural network architecture (i.e., model or individual) is encoded as a sequence of numbers. Each model is trained and the performance on a validation set is used as the fitness of the model. According to the fitness, reproductions and mutations are performed, and then, new high-performance “children”  (i.e., model) are generated. Recently, Liu et al. \cite{liu2018darts} proposed a method that transforms a discrete architecture space to a continuous space, which enables the gradient-based optimization method to learn for a suitable neural architecture. Due to the simplicity and time-saving property of the gradient-based method, gradient-based learning strategies have become popular. An attempt in remote sensing is that Chen et al. \cite{chen2019automatic} proposed an automatic CNN design approach in remote sensing HSI image classification.

\begin{figure*}[ht]
	\centering
	\includegraphics[width=6in]{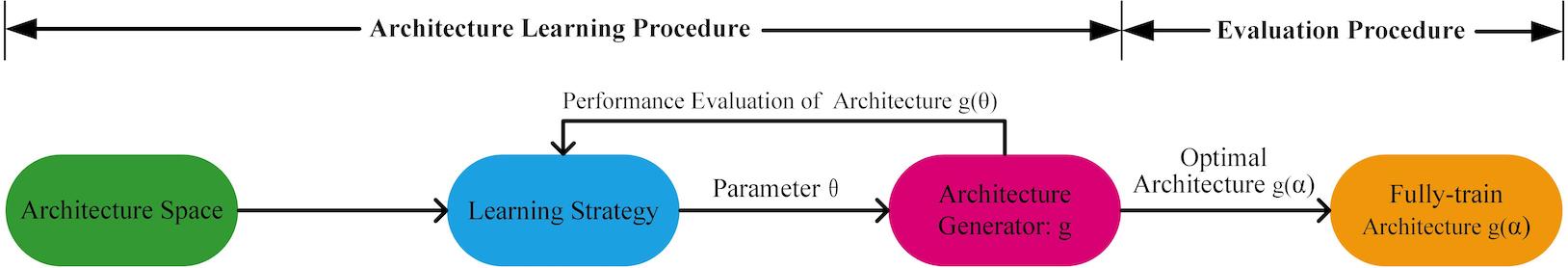}
	\caption{Illustration of architecture learning procedure of remote sensing scene classification.}
	\label{learning_proc}
\end{figure*}

\section{Method}
\subsection{Architecture learning procedure}
The available CNN architectures for remote sensing scene classification have been manually designed by experts in a trial and error way. In recent years, much progress has been achieved and CNN-based remote sensing scene classifiers have demonstrated a good classification performance. However, the manual design of a CNN classifier usually takes a long time. Moreover, deep architectures are data set dependent and they need to be
changed and adapted from one data set to another. Therefore, how to automatically design a proper CNN architecture is an important direction in the remote sensing scene classification.

Designing a suitable architecture can be seen as a learning procedure. Fig. \ref{learning_proc} shows the procedure of learning an architecture in an automatic way. There are three parts in the
architecture-learning system: Architecture space, learning strategy, and classification performance estimation. Architecture space is a collection of CNN architectures that
can be represented in principle. Appropriate Architecture space definition can reduce the size of architecture space and also leads to novel CNN architectures. The architecture space of our framework will be described at next section.The learning strategy is the core part of devising a well-performing CNN architecture. In general, it is a trade-off
between speed and performance. For the task of remote sensing scene classification, here, we use a single graphical processing unit (GPU) card to train the classifier. Therefore, a fast algorithm is important. Evaluation of architectures is the last part of the learning procedure in Fig. \ref{learning_proc}. In this paper, overall classification accuracy on validation samples is used to evaluate the performance of a specific CNN architecture.

\subsection{Architecture Space}

We define a cell to be a small fully convolutional module, typically repeated multiple times to form the entire neural network. More specifically, a cell is a directed acyclic graph consisting of B blocks. Each block is a two-branch structure, mapping from 2
input tensors to 1 output tensor. Block i in cell l may be specified using a 5-tuple $(I_1, I_2, O_1, O_2, C)$, where $I_1,I_2 \in I^l_i$ are selections of input tensors,$O_1, O_2 \in O$  are selections of layer types applied to the corresponding input tensor, and
$C \in C$ is the method used to combine the individual outputs of the two branches to form this block’s output tensor, $H_i^l$. The cell’s output tensor $H^l$ is simply the concatenation of the blocks’ output tensors $\left \{ H_1^l, ..., H_B^l \right\}$.
The set of possible input tensors, $I_i^l$, consists of the output of the previous cell $H^{l-1}$, the output of the previous previous cell $H^{l-2}$, and previous blocks’ output in the current cell $\left \{ H_1^l, ..., H_i^l \right\}$. Therefore, as we add more blocks in the cell, the next block has more choices as potential source of input.

\subsection{Learning Strategy}
In order to accelerate the learning procedure, gradient descent-based methods, which are useful in optimization problems (e.g., a neural network), are considered for the
learning procedure. Due to the continuity of the architecture space that is
required by the gradient descent-based method, the softmax operator is used over all possible operators to make the categorical choice of an operator continuous. Here, the output of an operator set is the weighted sum of the outputs of each operator in the operators set. 
\begin{equation}
\begin{aligned} c_{o}^{(i, j)} &=\frac{\exp \left(\alpha_{o}^{(i, j)}\right)}{\sum_{o^{\prime} \in O} \exp \left(\alpha_{o^{\prime}}^{(i, j)}\right)}  \end{aligned}
\end{equation}

\begin{equation}
\begin{aligned} o^{(i, j)}\left(n^{(j)}\right) &=\sum_{o \in O} c_{o}^{(i, j)} o\left(n^{(j)}\right)
\end{aligned}
\end{equation}
where $c_o^{(i, j)}$ is the coefficient of operation $o$ in the operation
set between node $n^{(i)}$ and $n^{(j)}$. The coefficients $c_o^{(i,j)}$ in
the operation set sum to 1 and are obtained by the softmax
operation. The coefficients $\alpha_o^{(i, j)}$ are randomly initialized and
optimized by the gradient descent method. Through the above described method, the architecture space changes from discrete to continuous, which can be solved using gradient descent. At the end of the learning procedure, the most likely operator,according to $o^{(i, j)} = argmax_{o\in O}\alpha _o^{i,j}$, is
selected as the final operator, and then, the network architecture is determined.

Analogous to the procedure of the manually designed neural architecture where the performance on the validation data set is used to guide the architecture design, Auto-CNN uses the gradient descent method to update the architecture parameter $\alpha$ based on the validation data set. The difference between
the two architecture design methods is that the former needs lots of prior experience on behalf of the analyst to adjust the neural structure, while the latter is fully automatic and data-dependent. Now, let $L_{train}$ and $L_{val}$ be the training and validation losses, respectively. The architecture optimization and the training of a CNN are a bi-level optimization problem with $\alpha$ as the architecture variables and $\omega$ as the weights in CNN.

\begin{equation}
\begin{array}{l}{\underset{\alpha}{\min} L_{\mathrm{val}}(w,\alpha)}\end{array}
\end{equation}

\begin{equation}
{\text { s.t. } w=\underset{w}{\operatorname{argmin}} L_{\mathrm{train}}(w, \alpha)}
\end{equation}

After the optimization, the obtained $\alpha ^*$ and $\omega ^*$ minimize the training and validation losses (i.e., $L_{train}$ and $L_{val}$), which are used for the architecture design and CNN, respectively. It can be summarized into the following two steps:

1. Update network weights $\omega$ by $\nabla_\omega L_{\mathrm{train}}(w,\alpha)$

2. Update network architecture $\alpha$ by $\nabla_\alpha L_{\mathrm{train}}(w,\alpha)$

When the architecture search process is finished, only one most likely operator on the connection between two nodes is preserved. Moreover, the output of each node is computed based on only the two strongest previous nodes. Here, the strength of the connection between two nodes is defined by the coefficient $\alpha _o^{(i,j)}$. Then, the architecture of Architecture-Learning CNN is determined. We train the Architecture-Learning CNN from scratch based on the training data set.

\subsection{Architecture Generator}
\begin{figure}[ht]
	\centering
	\includegraphics[width=3.4in]{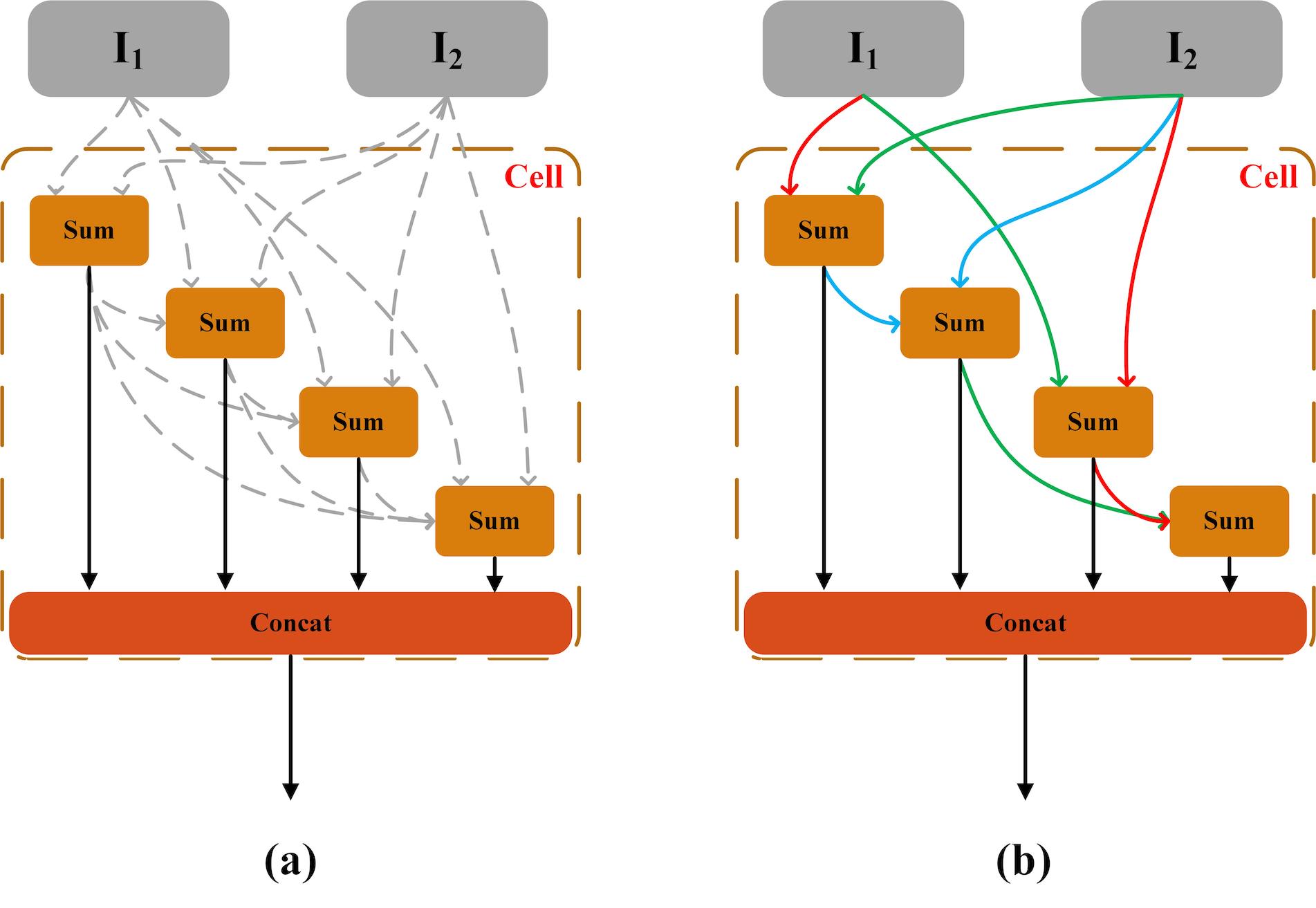}
	\caption{Computing Cell Architecture Space}
	\label{Arch_space}
\end{figure}

\begin{figure}[ht]
	\centering
	\includegraphics[width=1in]{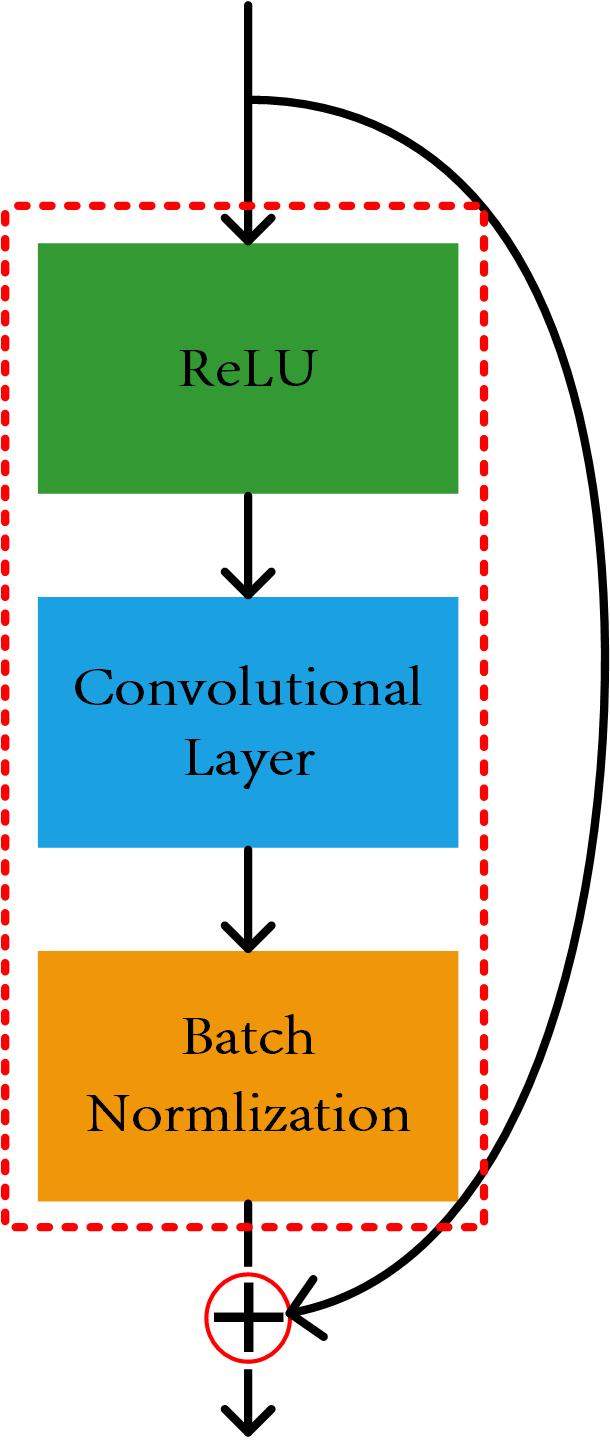}
	\caption{Residual convolution operator triplet.}
	\label{Op_triplet}
\end{figure}

Architecture Generator maps the set of parameters obtained by learning strategy to a computing cell shown as Fig. \ref{Arch_space}. 
 The set of possible layer types, $O$, consists of 7 operators(3 $\times$ 3 depthwise-separable conv, 5 $\times$ 5 depthwise-separable conv, 3 $\times$ 3 atrous conv with rate 2,  5 $\times$ 5 atrous conv with rate 2, 3 $\times$ 3 average pooling, 3 $\times$ 3 max pooling, skip connection) which all prevalent in modern CNNs. In addition, all the convolution operators are replaced by a triple illustrated as Fig. \ref{Op_triplet}.

\begin{figure*}[ht]
	\centering
	\includegraphics[width=6in]{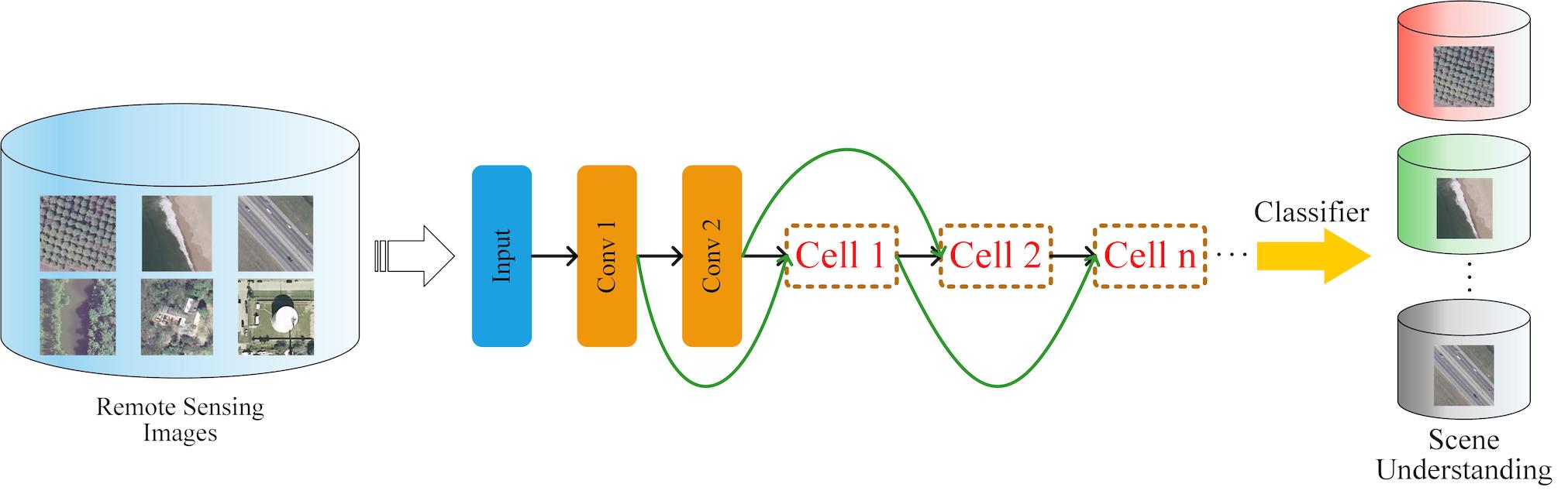}
	\caption{Our framework in remote sensing scene classification}
	\label{framework}
\end{figure*}
\section{Experiments}
To confirm the effectiveness of our proposed architecture learning procedure, we searched optimal architecture on three remote sensing scene classification data sets respectively and validated their performances. We will train architectures learned by our method and several classical CNN backbones with same parameter setting in our experiments, and compare the performance with our evaluation metrics. In addition, an ablation study is conducted to evaluate the effectiveness of atrous convolution in the architecture.
\subsection{Data Description}
In this section, we will choose three widely used remote sensing scene classification data sets (UC Merced Land-Use \cite{yang2010bag}, AID \cite{xia2017aid}, and NWPU-RESISC45 \cite{cheng2017remote}) to test the robustness and effectiveness of our proposed method. 
\subsubsection{UC Merced Land-Use Data Set}
The UC Merced Land-Use dataset is composed of 2100 aerial scene images divided into 21 land use scene classes. Each class contains 100 images with size of 256 $\times$ 256 pixels with a pixel spatial resolution of 0.3 m in the red green blue (RGB) color space. These images were selected from aerial orthoimagery downloaded from the United States Geological Survey (USGS) National Map of the following US regions: Birmingham, Boston, Buffalo, Columbus, Dallas, Harrisburg, Houston, Jacksonville, Las Vegas, Los Angeles, Miami, Napa, New York, Reno, San Diego, Santa Barbara, Seattle, Tampa, Tucson, and Ventura. It is not only the diversity of land-use categories contained in the dataset that makes it challenging. Some highly overlapped classes such as dense residential, medium residential and sparse residential are included in this dataset, which are mainly different in the density of structures and makes the dataset more difficult to classify. This dataset has been widely used for the task of remote sensing image scene classification.
\subsubsection{Aerial Image Data Set}
AID is large-scale aerial image dataset, which was collected from Google Earth imagery and is a more challenging dataset compared with The UC Merced Land-Use dataset because of The following reasons. First, the AID dataset contains more scene types and images. In detail, it has 10,000 images with a fixed size of 600 $\times$ 600 pixels within 30 classes. Some similar classes make the interclass dissimilarity smaller, and the number of images of different
scene types differs from 220 to 420. Moreover, AID images were chosen under different times and seasons and different imaging conditions, and from different countries And regions around the world. Finally, AID images have the property of multiresolution, changing from approximately 8 m to about half a meter.
\subsubsection{NWPU-RESISC45 Data Set}
NWPU-RESISC45 dataset is more complex than UC Merced Land-Use and AID datasets and
consists of a total of 31,500 remote sensing images divided into 45 scene classes. Each class includes 700 images with a size of 256 $\times$ 256 pixels in the RGB color space. This dataset was extracted from Google Earth by the experts in the field of remote sensing image interpretation. The spatial resolution varies from approximately 30 to 0.2 m per pixel. This dataset covers more than 100 countries and regions all over the world with developing, transitional, and highly developed economies.

\subsection{Implementation Details}
In this section, we will explain the parameter setting for our experiments. Our experiments can be divided into two primary parts, an architecture learning part and an architecture evaluating part. Firstly, we search the optimal architecture through our architecture learning procedure, and then we evaulate the learned optimal task-special architecture on corresponding datasets and compare the performance with classical CNN models under the same training settings to explain the effectiveness of our method. In the architecture learning procedure, we divide a data set into two isometrical parts, i.e. training set and validation set. Training set is used to train our learning strategy, validation set is used to evaulate the performances of architecture. And the architecture with best performance on the validation set will be chosen to be the optimal architecture. And in the architecture evaluation, the division of data sets depends on the setting of experiments. 

1) \emph{Architecture Space:} The operators set $O$ used in our architecture space contains 3 $\times$ 3 and 5 $\times$ 5 separable convolution residual triplet, 3 $\times$ 3 and 5 $\times$ 5 atrous convolution residual triplet, 3 $\times$ 3 max pooling, 3 $\times$ 3 average pooling and identity. It is worth noting that atrous convolution\cite{chen2017deeplab} (the same as dilated convolution\cite{yu2015multi}) which is merely used in remote sensing scene classification task appears in our operators set for its potential to abtain larger context. Every set of parameters in the architecture space can be map to a computing cell. Illustrated as Fig.4 (b), the input of $n$ th cell is the output of $n-1$ th cell and the output of $n-2$ th cell. For an acceptable computing time of the architecture learning procedure, we set the number of weighted sum node to 4, and every weighted sum node only aggregates 2 output from previous nodes. The output of these four weighted sum nodes are concat\cite{szegedy2015going} together as the output of cell. And we search two types of computing cells\cite{zoph2018learning} for a better performance in our exprtiments. A normal cell, the output resolution of feature map will stay the same with the input resolution for the stride of all oprators is set to 1, and a reduce cell, the stride of all oprators is set to 2. And the number of reduce cell used in our architecture is set to 2 in our experiments.

2) \emph{Architecture Learning Procedure:} As described in section III, when the learning strategy find a set of parameters, architecture generator will map it to a cell, and then stack the cell one by one. As illustrated in Fig. \ref{framework}, the input data pass two convolution layers (Conv1 is 3 $\times$ 3 with stride 2, and Conv2 is 3 $\times$ 3 with stride 1) in sequence, and normal cell and reduce cell are stacked by several times to complete the CNN. We tried multiple times to obtain the best number of cells for every dataset, i.e., the number of cells is set to 6 for UC Merced Land-Use data set, 7 for AID and 10 for NWPU-RESISC45. Then we train the CNN and architecture parameters alternatingly from scratch based on Pytorch with the NVIDIA 1080Ti. We resize the remote sensing images into 32 $\times$ 32 for the heavily memory consuming of architecture learning process. The training parameters are as follow. We train our CNN models using stochastic gradient descent, and set the batch size to 64 with the initial learning rate 0.025, momentum of 0.9 and a weight decay of $3 \times 10^{-4}$ for 50 epochs. And we train our architecture parameters with the learning rate of $3 \times 10^{-4}$ and weight decay $1 \times 10^{-3}$. 
\begin{figure*}[ht]
	\centering
	\includegraphics[width=6in]{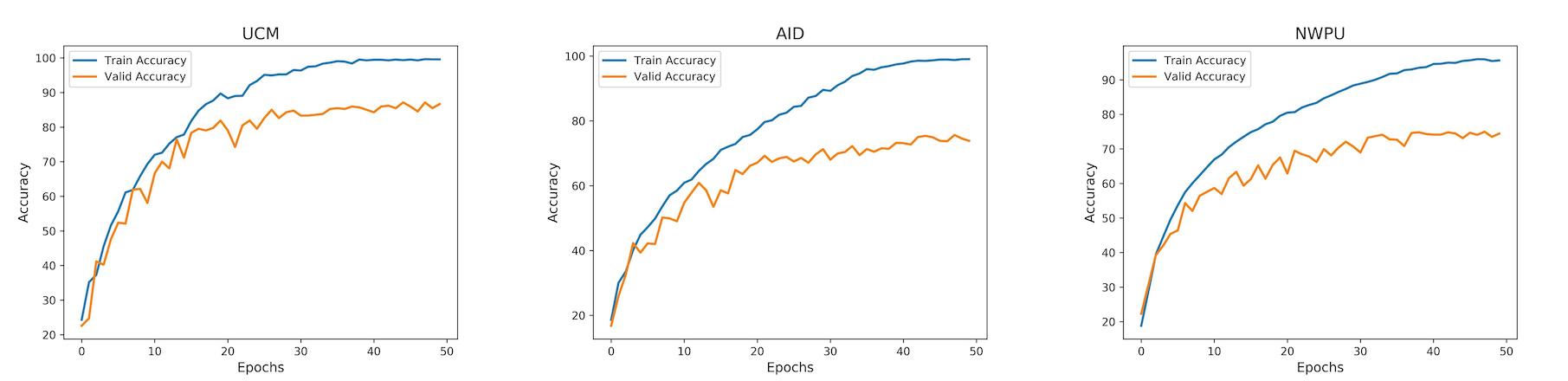}
	\caption{The train snd valid Accuracy during architecture learning procedure}
	\label{search}
\end{figure*}

\begin{figure*}[!h]
	\centering
	\includegraphics[width=4in]{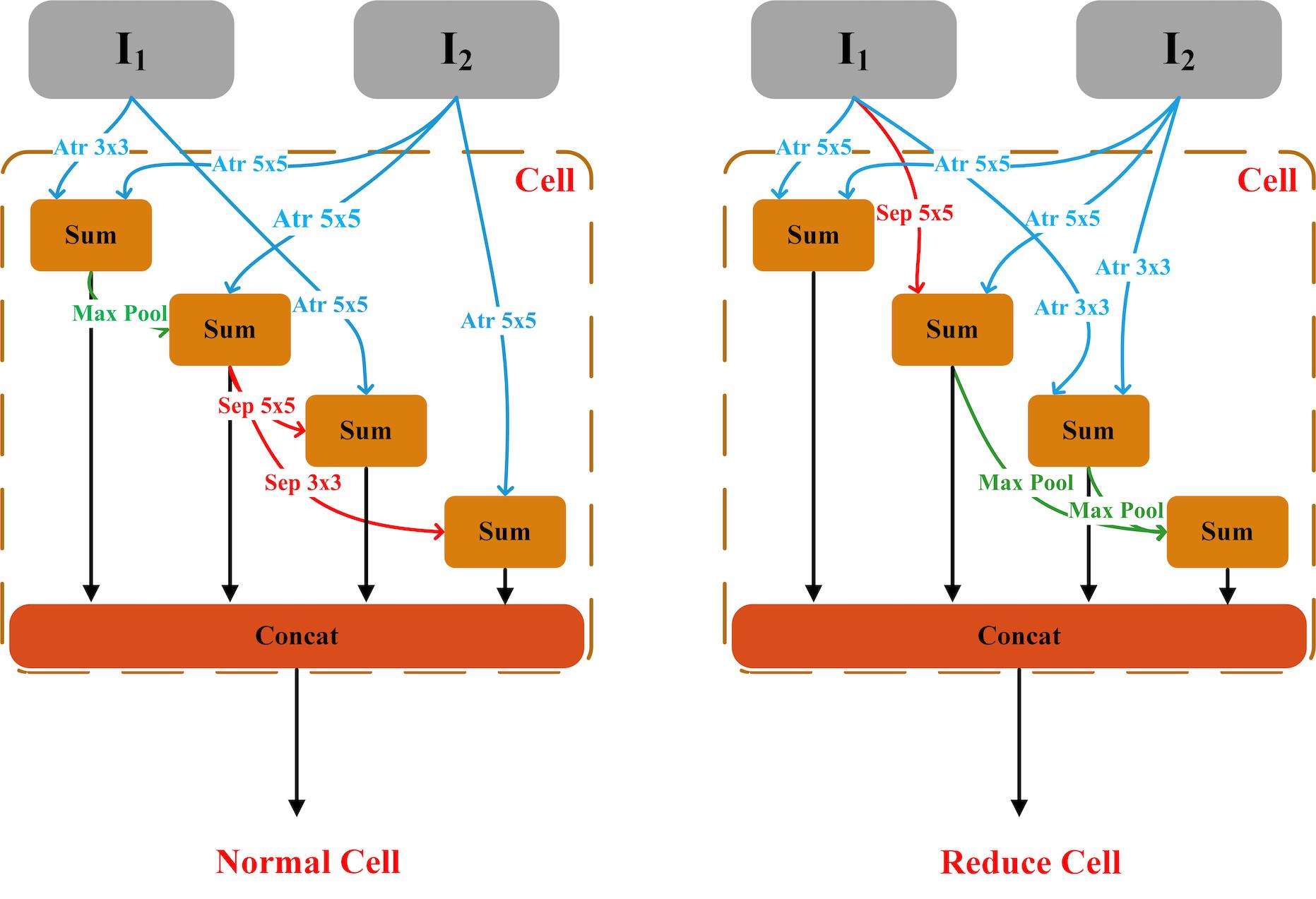}
	\caption{The optimal cell on the UCM data set}
	\label{cell_ucm}
\end{figure*}

\begin{figure*}[!h]
	\centering
	\includegraphics[width=4in]{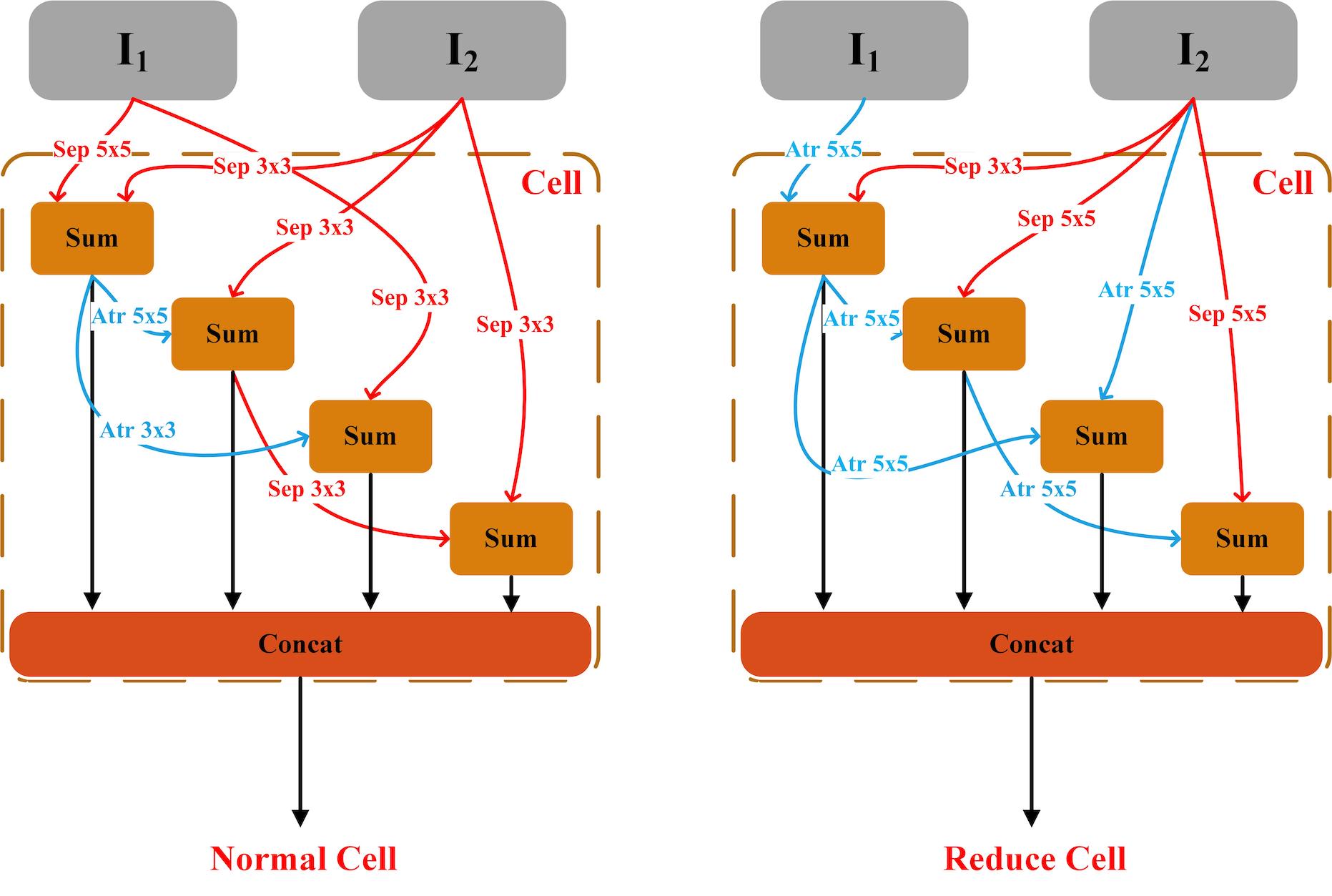}
	\caption{The optimal cell on the AID}
	\label{cell_aid}
\end{figure*}

\begin{figure*}[ht]
	\centering
	\includegraphics[width=4in]{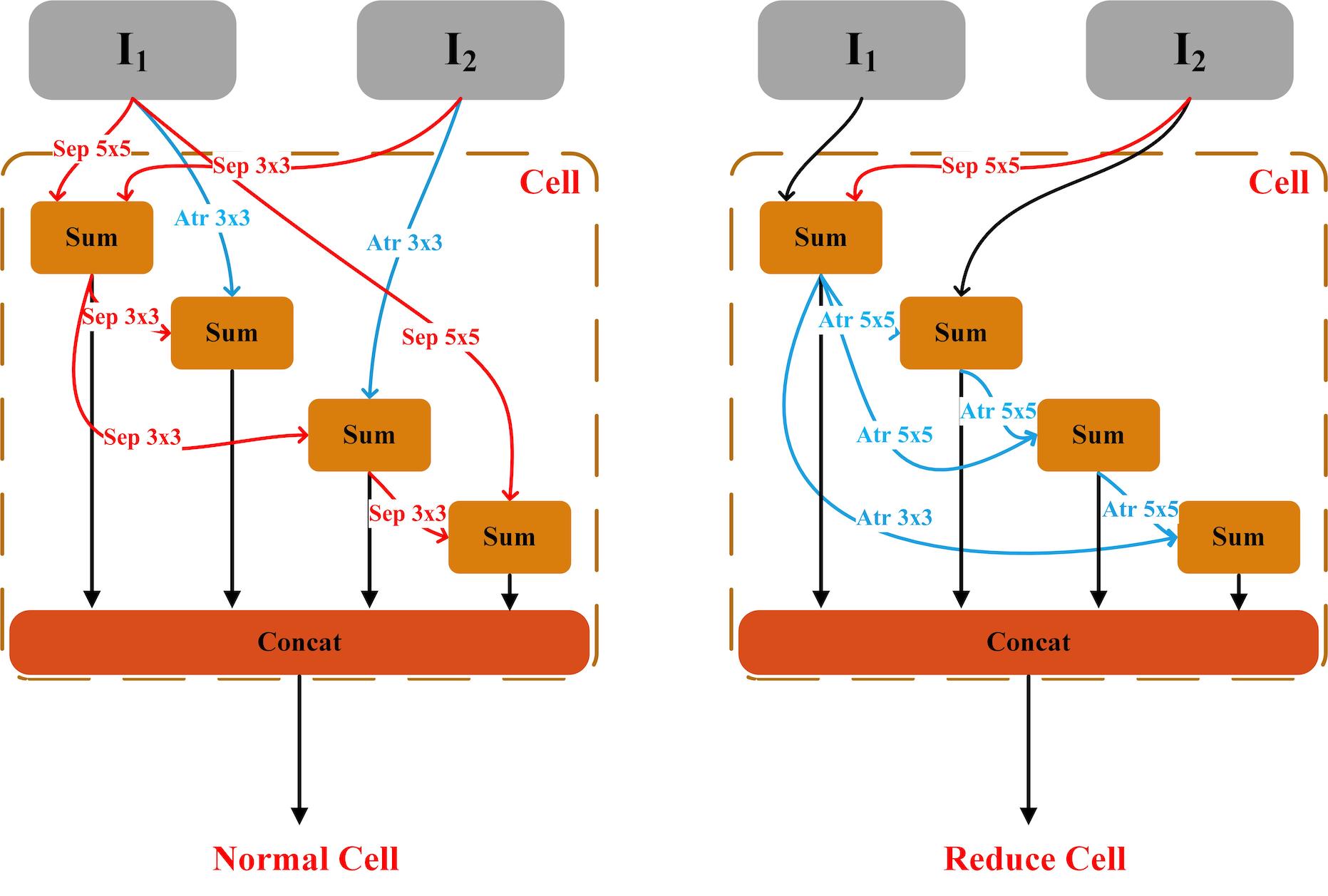}
	\caption{The optimal cell on the NWPU-RESISC45 data set}
	\label{cell_nwpu}
\end{figure*}

3) \emph{Evaluation Procedure:} In the architecture evaluation procedure, we train architectures learned by our method and classical CNN architectures with the same settings. The optimal cells are stacked the same times as the architecture learning procedure. Images are all resized to 224 $\times$ 224 with random horizontal flip before inputting CNNs. Our architectures are full-trained\cite{nogueira2017towards} with a batch size of 64 using stochastic gradient descent, initial learning rate of 0.1 with a decay of 0.97 in every epoch, momentum of 0.9, and every model is trained for 150 epochs.

\subsection{Experiment Results}
In this section, we will demonstrate the architectures learned by our method and compare the performances of these architectures with some classical CNN architectures being widely used in remote sensing scene classification. For our datasets have the similar numbers of images in each classes, average accuracy (AA) will close to overall accuracy (OA). Thus we exclude it from the evaluation metrics in our results. OA and confusion matrix(CM) are used for declaring the effectiveness of our method.

1) \emph{Architecture Learning Procedure:} We apply our architecture learning procedure on the three data sets. As it shown in Fig. \ref{search}, we learn the optimal architectures for 50 epochs. During the architecture learning process, the training accuracy and valid accuracy continue to increase and finally approach convergence at the 50-th epoch. And we also obtian the optimal cells as shown in Fig. \ref{cell_ucm}-\ref{cell_nwpu}. The sum in pictures represent weighted sum operation. Sep 3 $\times$ 3 and 5 $\times$ 5 represent 3 $\times$ 3 and 5 $\times$ 5 separable convolution residual triplet. Atr 3 $\times$ 3 and 5 $\times$ 5 represent 3 $\times$ 3 and 5 $\times$ 5 atrous convolution residual triplet. Max Pool means 3 $\times$ 3 max pooling, and the black line without any text represent identity. We can find that all of the cells contain separable convolution and atrous convolution. Atrous convolution is known as its larger field of view of filters to catch larger context. Architectures with atrous convolution may get better performance by meeting the demand for remote sensing scene classification with larger context.
 
2) \emph{Architecture Evaluation Procedure:}  A performance comparison between the optimal architectures we learned and eight classical CNN architectures is performed, such as AlexNet\cite{krizhevsky2012imagenet}, VGG16\cite{simonyan2014very} and ResNet-50\cite{he2016deep}. We compare the numbers of parameters between these architectures, shown in Table \ref{table_para}. For the optimal cells are different between datasets, and cells are stacked in different times, the parameters of our optimal architectures are different. It is easy to observe that the architectures we found have smaller numbers of parameters than the calssical ones. We train all the architectures in a full-trained strategy. Comparing with the fine-tuned strategy, this strategy is pruned to overfitting and will result in a performance decrease\cite{nogueira2017towards}.

\begin{figure}[!ht]
	\centering
	\includegraphics[width=3.4in]{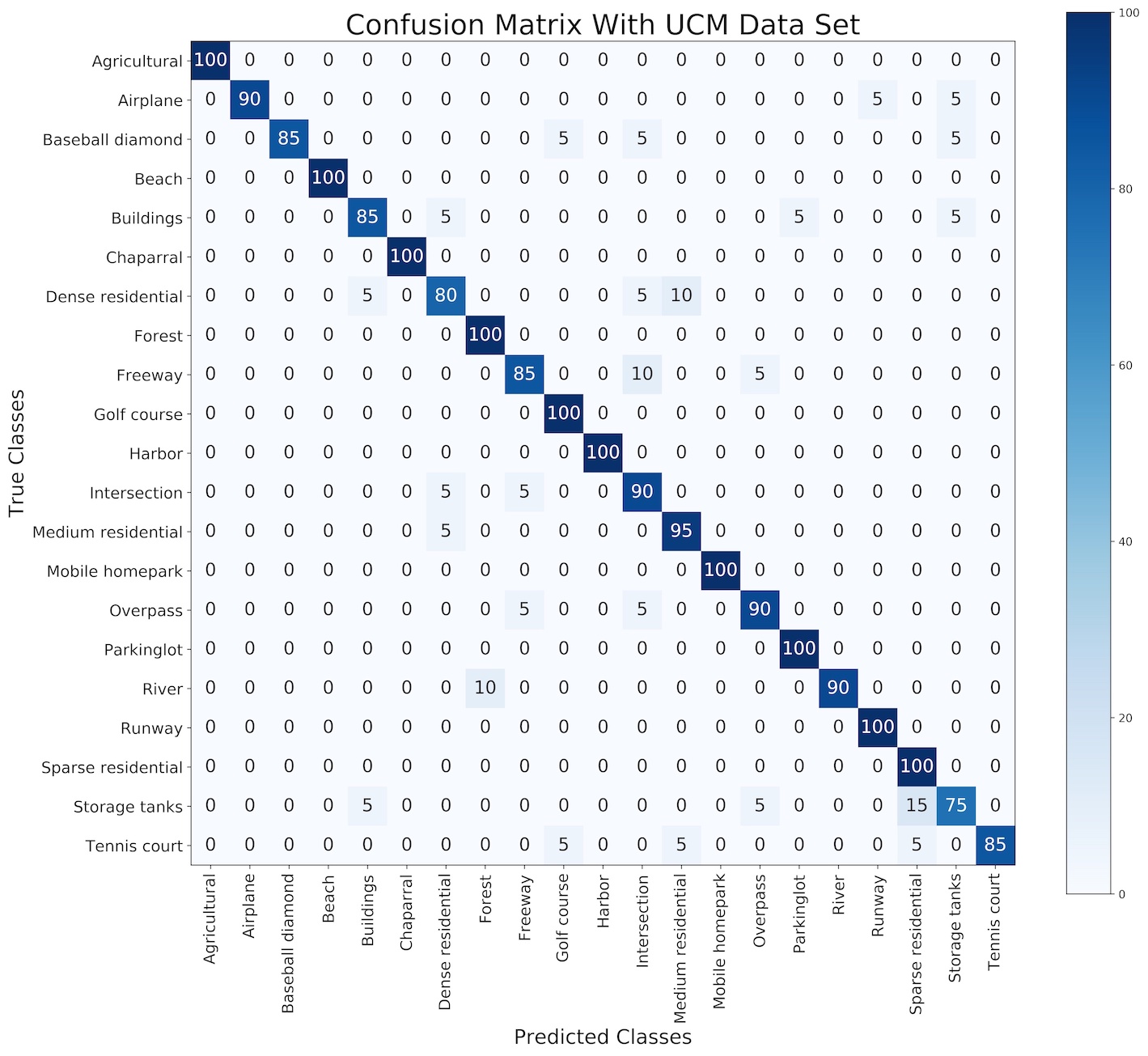}
	\caption{CM on UCM under the training ratio of 50\%}
	\label{cm_ucm}
\end{figure}

For UC Merced Land-Use data set, we stack the learned optimal cells in 6 times, and train the architecture from scratch for 150 epochs. We train the dataset in two training ratio. The result is shown in Table \ref{table_all}. Our method get the best performance in both the training ratio of 50\% and 80 \%. 
We also make a CM to further analyze the effect of the architecture learned by our architecture learning procedure, as shown in Fig. \ref{cm_ucm}. 
It can be observed that the performance using 80\% images for training is lower than using 50\% images for training. The reason is that there are only 2100 images in UCM data set. More training data may destroy the robustness of the model. 

\begin{table*}[!ht]
	\renewcommand{\arraystretch}{1.3}
	\caption{Parameters comparison between architectures in our experiments}
	\label{table_para}
	\centering
	\begin{tabular}{ccc}
		\hline
		Architecture & \#parameters(millions) \\
		%& 80\%  images for training \\
		\hline
		%\quad & OA (\%) & Kappa & OA (\%) & Kappa \\
		AlexNet\cite{krizhevsky2012imagenet} & 60 \\
		% & 82.852  \\
		VGG16\cite{simonyan2014very} & 138\\
		Googlenet\cite{szegedy2015going} & 6.8\\
		InceptionV3\cite{szegedy2016rethinking} & 23.2 \\
		% & 83.810 \\
		ResNet-50\cite{he2016deep} & 25.6 \\
		%& 88.810 \\.
		DenseNet\cite{huang2017densely} & 13.7\\
		Wide ResNet\cite{zagoruyko2016wide} & 68.9\\
		MobileNetV2\cite{sandler2018mobilenetv2} & 6.9\\
		Optimal Architecture on UCM  & 2.626 \\
		Optimal Architecture on AID  & 3.745 \\
		Optimal Architecture on NWPU  & 5.180 \\
		%& 89.524\\
		\hline
	\end{tabular}
\end{table*}

\begin{figure*}[!ht]
	\centering
	\includegraphics[width=4.5in]{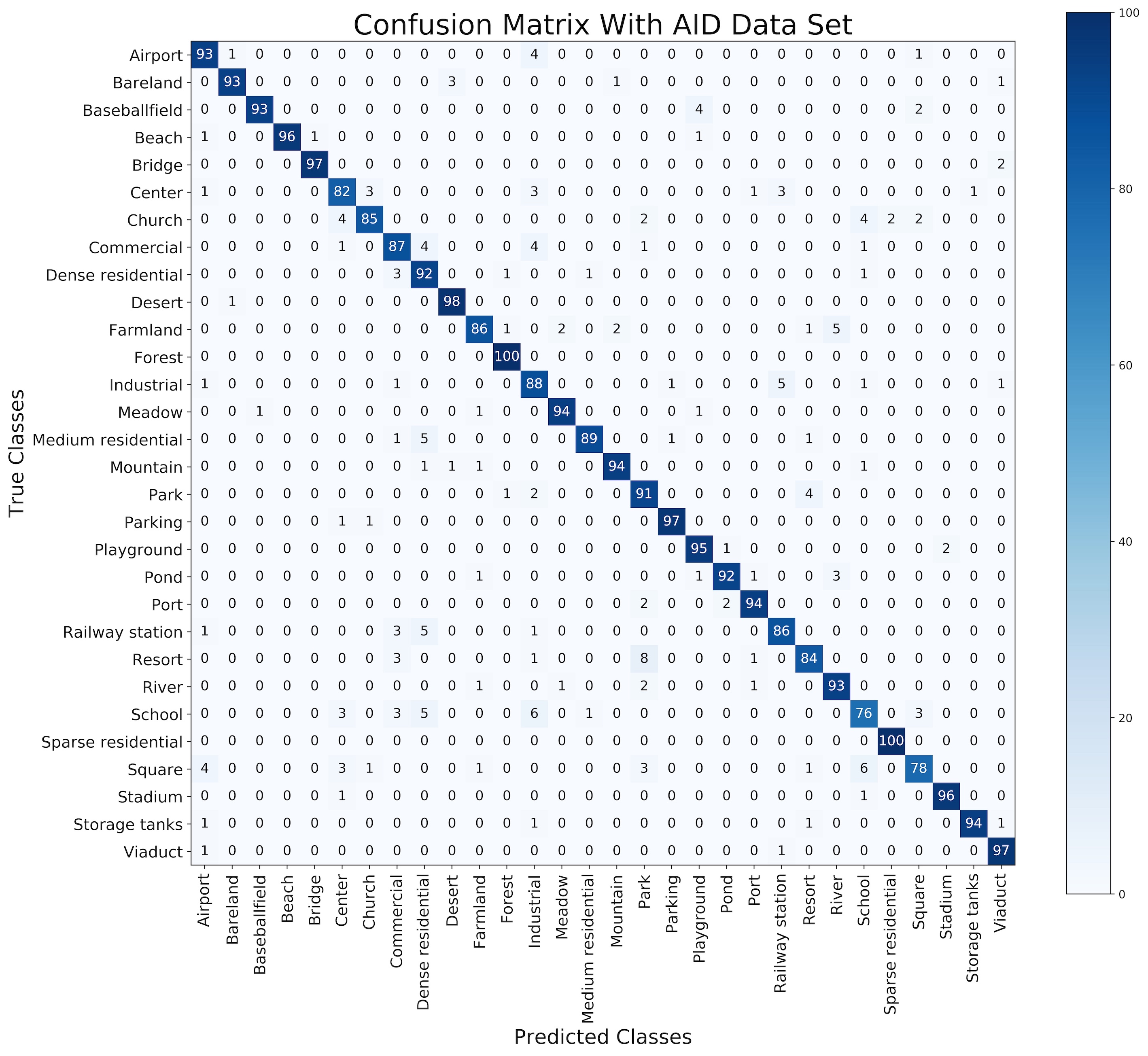}
	\caption{CM on AID under the training ratio of 80\%}
	\label{cm_aid}
\end{figure*}

For AID, we stack the learned optimal cells in 7 times, and train the architecture from scratch for 150 epochs. We train the dataset in two training ratio. The result is shown in Table \ref{table_all}. Our method get the best performance in both the training ratio of 50\% and 80 \%. 
We also make a CM to further analyze the effect of the architecture learned by our architecture learning procedure, as shown in Fig. \ref{cm_aid}.

\begin{figure*}[!ht]
	\centering
	\includegraphics[width=4.5in]{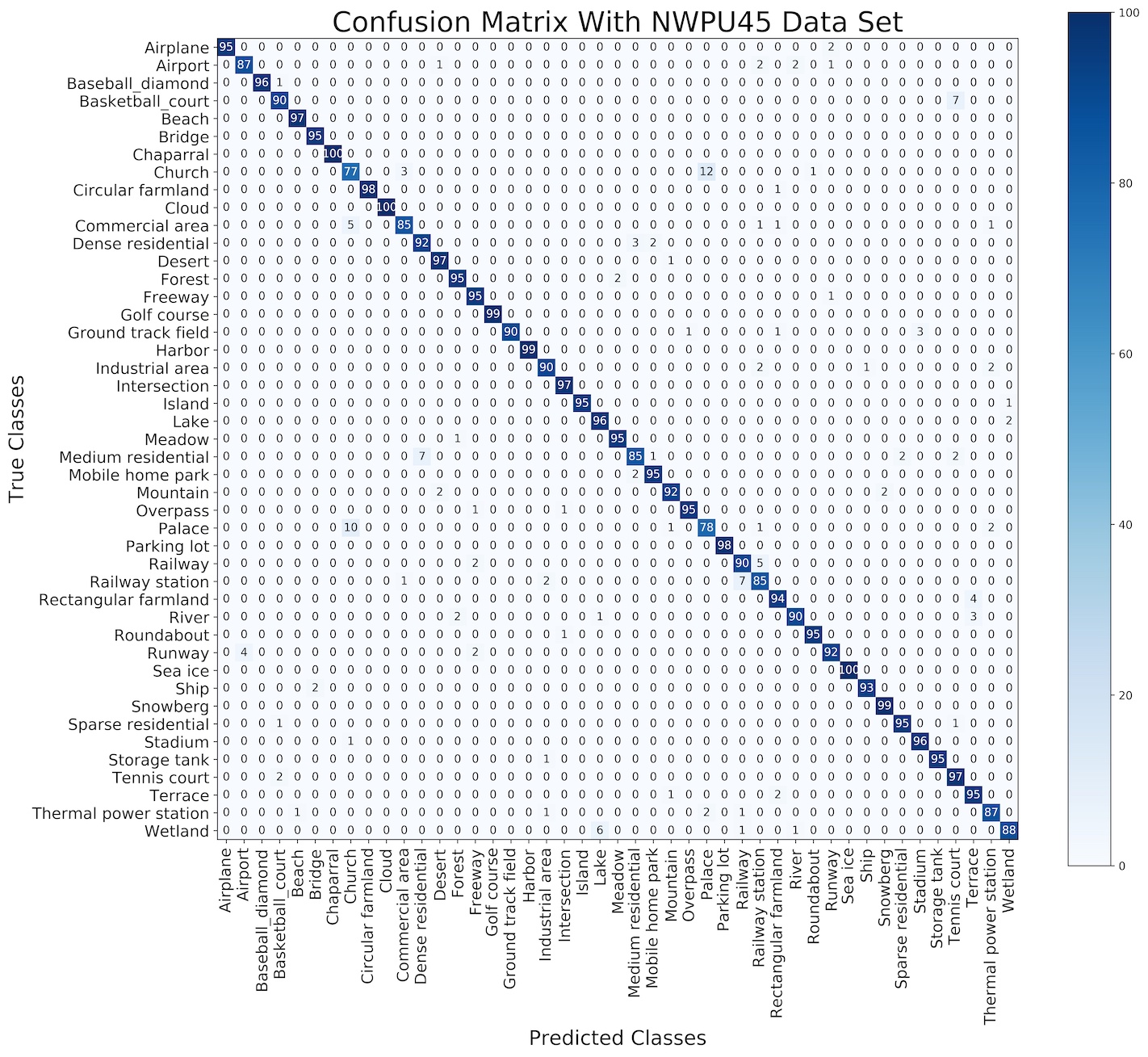}
	\caption{CM on NWPU under the training ratio of 80\%}
	\label{cm_nwpu}
\end{figure*}

For NWPU, we stack the learned optimal cells in 10 times, and train the architecture from scratch for 150 epochs. We train the dataset in two training ratio. The result is shown in Table \ref{table_all}. Our method get the best performance in both the training ratio of 50\% and 80 \%. 
We also make a CM to further analyze the effect of the architecture learned by our architecture learning procedure, as shown in Fig. \ref{cm_nwpu}

\begin{figure*}[ht]
	\centering
	\includegraphics[width=4in]{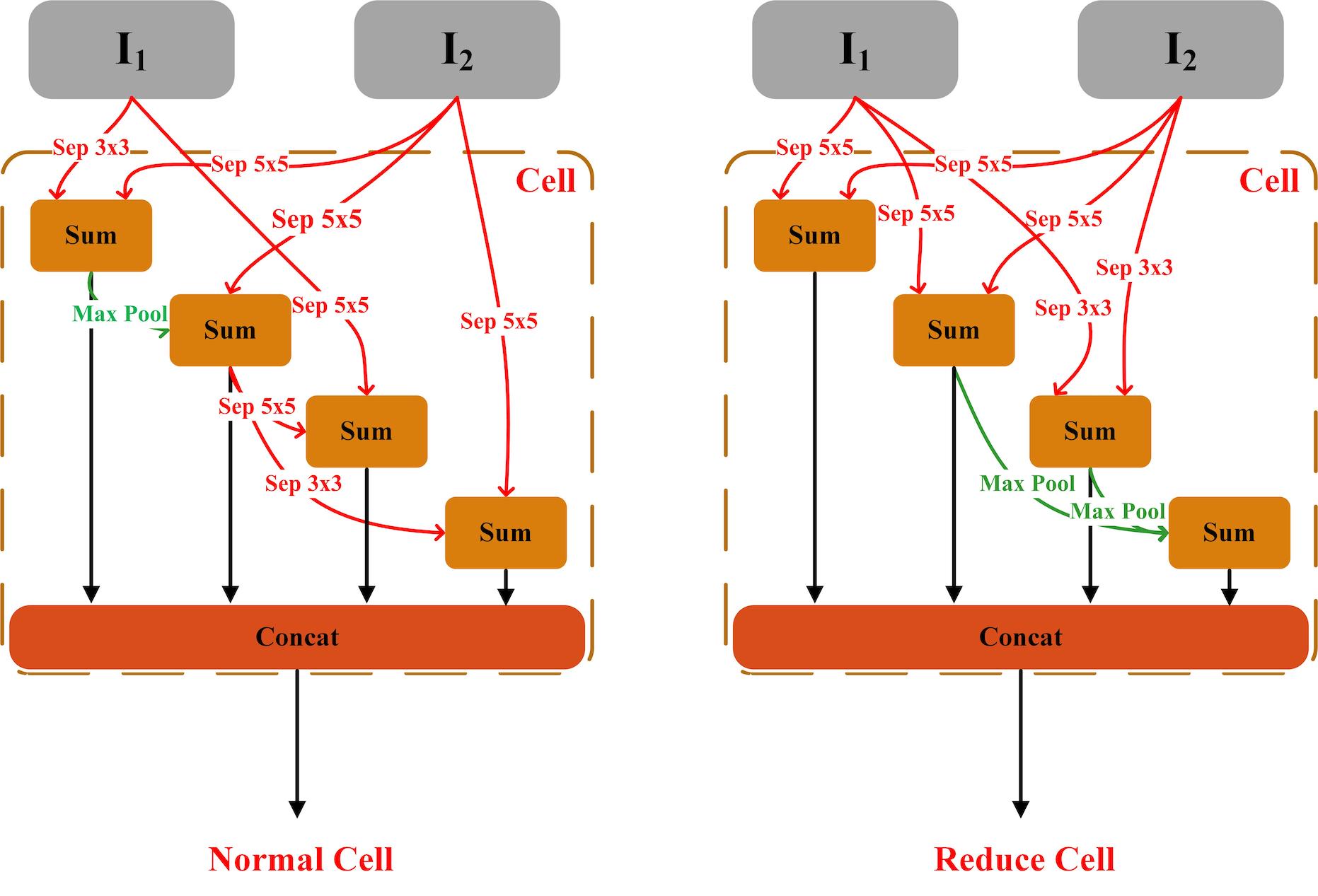}
	\caption{The optimal cell on the UCM in ablation study}
	\label{cell_ab}
\end{figure*}

\begin{table*}[!ht]
	\renewcommand{\arraystretch}{1.3}
	\caption{Comparison between methods}
	\label{table_all}
	\centering
	\begin{tabular}{c|c|c|c|c|c|c}
		\hline
		\quad & \multicolumn{2}{c|}{UCM} & \multicolumn{2}{c|}{AID} &\multicolumn{2}{c}{NWPU}\\
		\hline
		\multirow{2}{*}{Methods} & \multicolumn{2}{c|}{TP\%} & \multicolumn{2}{c|}{TP\%} & \multicolumn{2}{c}{TP\%}\\
		\cline{2-7} & 50\%  & 80\%  & 50\% & 80\%  & 50\% & 80\%  \\
		\hline
		AlexNet & 77.676 $\pm$ 1.183  & 88.619 $\pm$ 0.200 & 79.724 $\pm$ 0.514 & 84.900 $\pm$ 0.382 & 82.697 $\pm$ 0.069 & 86.321 $\pm$ 0.281\\
		VGG16 & 81.467 $\pm$ 0.306 & 88.953 $\pm$ 0.272 & 82.416 $\pm$ 0.506 & 87.760 $\pm$ 0.213 & 85.003 $\pm$ 0.224 & 89.502 $\pm$ 0.150\\
		Googlenet & 82.705 $\pm$ 0.987 & 87.666 $\pm$ 1.070 & 82.772 $\pm$ 1.537 & 88.150 $\pm$ 0.777 & 86.592 $\pm$ 0.824 & 91.613 $\pm$ 0.657\\
		InceptionV3 & 81.676 $\pm$ 1.228 & 87.762 $\pm$ 1.364 & 82.444 $\pm$ 0.745 & 89.030 $\pm$ 1.272 & 88.305 $\pm$ 0.479 & 92.584 $\pm$ 0.267\\
		ResNet-50 & 83.371 $\pm$ 1.104 & 88.714 $\pm$ 1.755 & 82.476 $\pm$ 0.683 & 88.100 $\pm$ 0.651 & 87.774 $\pm$ 0.871 & 92.013 $\pm$ 0.291\\
		DenseNet & 83.790 $\pm$ 0.611 & 89.667 $\pm$ 0.916 & 82.760 $\pm$ 0.703 & 87.500 $\pm$ 0.242 & 88.692 $\pm$ 0.517 & 92.708 $\pm$ 0.173\\
		Wide ResNet & 82.476 $\pm$ 0.741 & 88.857 $\pm$ 0.488 & 82.160 $\pm$ 0.467 & 88.250 $\pm$ 0.941 & 87.478 $\pm$ 0.967 & 92.308 $\pm$ 0.171\\
		MobileNetV2 & 84.000 $\pm$ 0.415 & 90.000 $\pm$ 0.996 & 85.680 $\pm$ 0.331 & 89.330 $\pm$ 0.372 & 89.543 $\pm$ 0.224 & 92.289 $\pm$ 0.190\\
		Our Methods & \textbf{88.628 $\pm$ 1.134} & \textbf{93.428 $\pm$ 0.706} & \textbf{86.996 $\pm$ 0.750} & \textbf{91.120 $\pm$ 0.498} & \textbf{89.725 $\pm$ 0.386} & \textbf{93.667 $\pm$ 0.241}\\
		\hline
	\end{tabular}
\end{table*}

\subsection{Ablation Study}
For the observation that all the optimal architectures found by our architecture learning procedure contain atrous convolution operator, the larger context caught by atrous convolution may crucial for remote sensing scene classification. To further vertify our conclusion, we design an ablation study.

We perform ablation experiments in two ways. One is to replace all the atrous convolutions in our optimal cells by separable convolution with the same size, as the example we do for the optimal cell of UCM illustrated in Fig. \ref{cell_ab}. The other is to exclude atrous convolutions in the architecture learning procedure, and then repeat the previous architecture learning procedure to observe the performance of the optimal architectures.
We train these architectures with the same training settings described in Section IV.B.
The performance comparison between the architectures with and without atrous convolution is shown in Table \ref{table_ab}. 
Obviously, all the architectures without atrous convolution can be observed a performance decrease from 3.16\% to 0.62\%. It is indicated that the architectures with atrous convolution can learn a better representation than without it.

\begin{table*}[!t]
	\renewcommand{\arraystretch}{1.3}
	\caption{Performances comparison between the model with and without atrous convolution}
	\label{table_ab}
	\centering
	\begin{tabular}{c|c|c|c|c|c|c}
		\hline
		\quad & \multicolumn{2}{c|}{UCM} & \multicolumn{2}{c|}{AID} &\multicolumn{2}{c}{NWPU}\\
		\hline
		\multirow{2}{*}{Methods} & \multicolumn{2}{c|}{TP\%} & \multicolumn{2}{c|}{TP\%} & \multicolumn{2}{c}{TP\%}\\
		\cline{2-7} & 50\%  & 80\%  & 50\% & 80\%  & 50\% & 80\%  \\
		\hline
		Our Methods & \textbf{88.628 $\pm$ 1.134} & \textbf{93.428 $\pm$ 0.706} & \textbf{86.996 $\pm$ 0.750} & \textbf{91.120 $\pm$ 0.498} & \textbf{89.725 $\pm$ 0.386} & \textbf{93.667 $\pm$ 0.241}\\
		Ablation Set 1 & 87.001 $\pm$ 0.465 & 91.571 $\pm$ 0.574 & 86.200 $\pm$ 0.821 & 91.040 $\pm$ 0.947 & 89.351 $\pm$ 0.328 & 93.271 $\pm$ 0.562\\
		Ablation Set 2 & 87.526 $\pm$ 0.343 & 92.754 $\pm$ 0.483 & 86.327 $\pm$ 0.643 & 90.872 $\pm$ 0.569 & 89.486 $\pm$ 0.223 & 93.197 $\pm$ 0.376\\
		\hline
	\end{tabular}
\end{table*}

\section{Conclusion}
In this paper, we proposed an architecture learning procedure which can learn a task-special CNN architecture for remote sensing scene classification. And experiment results indicate that the task-special CNN architecture outperform several classical human-designed architectures on three remote sensing scene classification benchmarks. 
Moreover, architecture determines the function of CNN, this architecture-learned paradigm may help to understand which representations are important for remote sensing scene classification tasks and lead to a better backbone for remote sensing scene classification.

% or


%\appendices

%\section{}

%\section*{Acknowledgment}


\ifCLASSOPTIONcaptionsoff
  \newpage
\fi



%\bibliographystyle{IEEEtran}
\newpage
\bibliographystyle{IEEEtran}
\bibliography{RsNas.bib}

\begin{thebibliography}{10}
\providecommand{\url}[1]{#1}
\csname url@samestyle\endcsname
\providecommand{\newblock}{\relax}
\providecommand{\bibinfo}[2]{#2}
\providecommand{\BIBentrySTDinterwordspacing}{\spaceskip=0pt\relax}
\providecommand{\BIBentryALTinterwordstretchfactor}{4}
\providecommand{\BIBentryALTinterwordspacing}{\spaceskip=\fontdimen2\font plus
\BIBentryALTinterwordstretchfactor\fontdimen3\font minus
  \fontdimen4\font\relax}
\providecommand{\BIBforeignlanguage}[2]{{%
\expandafter\ifx\csname l@#1\endcsname\relax
\typeout{** WARNING: IEEEtran.bst: No hyphenation pattern has been}%
\typeout{** loaded for the language `#1'. Using the pattern for}%
\typeout{** the default language instead.}%
\else
\language=\csname l@#1\endcsname
\fi
#2}}
\providecommand{\BIBdecl}{\relax}
\BIBdecl

\bibitem{Plaza2011Parallel}
A.~Plaza, J.~Plaza, A.~Paz, and S.~S{\'a}Nchez, ``Parallel hyperspectral image
  and signal processing,'' \emph{Signal Processing Magazine IEEE}, vol.~28,
  no.~3, pp. 119--126, 2011.

\bibitem{Cantalloube2013Airborne}
H.~M.~J. Cantalloube and C.~E. Nahum, ``Airborne sar-efficient signal
  processing for very high resolution,'' \emph{Proceedings of the IEEE}, vol.
  101, no.~3, pp. 784--797, 2013.

\bibitem{cheng2017remote}
G.~Cheng, J.~Han, and X.~Lu, ``Remote sensing image scene classification:
  Benchmark and state of the art,'' \emph{Proceedings of the IEEE}, vol. 105,
  no.~10, pp. 1865--1883, 2017.

\bibitem{cheng2014multi}
G.~Cheng, J.~Han, P.~Zhou, and L.~Guo, ``Multi-class geospatial object
  detection and geographic image classification based on collection of part
  detectors,'' \emph{ISPRS Journal of Photogrammetry and Remote Sensing},
  vol.~98, pp. 119--132, 2014.

\bibitem{wang2016three}
Y.~Wang, L.~Zhang, X.~Tong, L.~Zhang, Z.~Zhang, H.~Liu, X.~Xing, and P.~T.
  Mathiopoulos, ``A three-layered graph-based learning approach for remote
  sensing image retrieval,'' \emph{IEEE Transactions on Geoscience and Remote
  Sensing}, vol.~54, no.~10, pp. 6020--6034, 2016.

\bibitem{zhu2016bag}
Q.~Zhu, Y.~Zhong, B.~Zhao, G.-S. Xia, and L.~Zhang, ``Bag-of-visual-words scene
  classifier with local and global features for high spatial resolution remote
  sensing imagery,'' \emph{IEEE Geoscience and Remote Sensing Letters},
  vol.~13, no.~6, pp. 747--751, 2016.

\bibitem{estoque2015pixel}
R.~C. Estoque, Y.~Murayama, and C.~M. Akiyama, ``Pixel-based and object-based
  classifications using high-and medium-spatial-resolution imageries in the
  urban and suburban landscapes,'' \emph{Geocarto International}, vol.~30,
  no.~10, pp. 1113--1129, 2015.

\bibitem{blaschke2001s}
T.~Blaschke, ``What's wrong with pixels? some recent developments interfacing
  remote sensing and gis,'' \emph{GeoBIT/GIS}, vol.~6, pp. 12--17, 2001.

\bibitem{blaschke2010object}
------, ``Object based image analysis for remote sensing,'' \emph{ISPRS journal
  of photogrammetry and remote sensing}, vol.~65, no.~1, pp. 2--16, 2010.

\bibitem{blaschke2014geographic}
T.~Blaschke, G.~J. Hay, M.~Kelly, S.~Lang, P.~Hofmann, E.~Addink, R.~Q.
  Feitosa, F.~Van~der Meer, H.~Van~der Werff, F.~Van~Coillie \emph{et~al.},
  ``Geographic object-based image analysis--towards a new paradigm,''
  \emph{ISPRS journal of photogrammetry and remote sensing}, vol.~87, pp.
  180--191, 2014.

\bibitem{Oliva2001Modeling}
A.~Oliva and A.~Torralba, ``Modeling the shape of the scene: A holistic
  representation of the spatial envelope,'' \emph{International Journal of
  Computer Vision}, vol.~42, no.~3, pp. 145--175, 2001.

\bibitem{Lowe2004Distinctive}
D.~G. Lowe, ``Distinctive image features from scale-invariant keypoints,''
  \emph{International Journal of Computer Vision}, vol.~60, no.~2, pp. 91--110,
  2004.

\bibitem{Dalal2005Histograms}
N.~Dalal and B.~Triggs, ``Histograms of oriented gradients for human
  detection,'' in \emph{IEEE Computer Society Conference on Computer Vision \&
  Pattern Recognition}, 2005.

\bibitem{lecun2015deep}
Y.~LeCun, Y.~Bengio, and G.~Hinton, ``Deep learning,'' \emph{nature}, vol. 521,
  no. 7553, p. 436, 2015.

\bibitem{scott2017training}
G.~J. Scott, M.~R. England, W.~A. Starms, R.~A. Marcum, and C.~H. Davis,
  ``Training deep convolutional neural networks for land--cover classification
  of high-resolution imagery,'' \emph{IEEE Geoscience and Remote Sensing
  Letters}, vol.~14, no.~4, pp. 549--553, 2017.

\bibitem{chaib2017deep}
S.~Chaib, H.~Liu, Y.~Gu, and H.~Yao, ``Deep feature fusion for vhr remote
  sensing scene classification,'' \emph{IEEE Transactions on Geoscience and
  Remote Sensing}, vol.~55, no.~8, pp. 4775--4784, 2017.

\bibitem{li2017integrating}
E.~Li, J.~Xia, P.~Du, C.~Lin, and A.~Samat, ``Integrating multilayer features
  of convolutional neural networks for remote sensing scene classification,''
  \emph{IEEE Transactions on Geoscience and Remote Sensing}, vol.~55, no.~10,
  pp. 5653--5665, 2017.

\bibitem{paoletti2018new}
M.~Paoletti, J.~Haut, J.~Plaza, and A.~Plaza, ``A new deep convolutional neural
  network for fast hyperspectral image classification,'' \emph{ISPRS journal of
  photogrammetry and remote sensing}, vol. 145, pp. 120--147, 2018.

\bibitem{nogueira2017towards}
K.~Nogueira, O.~A. Penatti, and J.~A. dos Santos, ``Towards better exploiting
  convolutional neural networks for remote sensing scene classification,''
  \emph{Pattern Recognition}, vol.~61, pp. 539--556, 2017.

\bibitem{krizhevsky2012imagenet}
A.~Krizhevsky, I.~Sutskever, and G.~E. Hinton, ``Imagenet classification with
  deep convolutional neural networks,'' in \emph{Advances in neural information
  processing systems}, 2012, pp. 1097--1105.

\bibitem{simonyan2014very}
K.~Simonyan and A.~Zisserman, ``Very deep convolutional networks for
  large-scale image recognition,'' \emph{arXiv preprint arXiv:1409.1556}, 2014.

\bibitem{yang2010bag}
Y.~Yang and S.~Newsam, ``Bag-of-visual-words and spatial extensions for
  land-use classification,'' in \emph{Proceedings of the 18th SIGSPATIAL
  international conference on advances in geographic information
  systems}.\hskip 1em plus 0.5em minus 0.4em\relax ACM, 2010, pp. 270--279.

\bibitem{penatti2015deep}
O.~A. Penatti, K.~Nogueira, and J.~A. Dos~Santos, ``Do deep features generalize
  from everyday objects to remote sensing and aerial scenes domains?'' in
  \emph{Proceedings of the IEEE conference on computer vision and pattern
  recognition workshops}, 2015, pp. 44--51.

\bibitem{chen2019automatic}
Y.~Chen, K.~Zhu, L.~Zhu, X.~He, P.~Ghamisi, and J.~A. Benediktsson, ``Automatic
  design of convolutional neural network for hyperspectral image
  classification,'' \emph{IEEE Transactions on Geoscience and Remote Sensing},
  2019.

\bibitem{bhagavathy2006modeling}
S.~Bhagavathy and B.~S. Manjunath, ``Modeling and detection of geospatial
  objects using texture motifs,'' \emph{IEEE Transactions on Geoscience and
  Remote Sensing}, vol.~44, no.~12, pp. 3706--3715, 2006.

\bibitem{dos2010evaluating}
J.~A. dos Santos, O.~A.~B. Penatti, and R.~da~Silva~Torres, ``Evaluating the
  potential of texture and color descriptors for remote sensing image retrieval
  and classification.'' in \emph{VISAPP (2)}, 2010, pp. 203--208.

\bibitem{yang2012geographic}
Y.~Yang and S.~Newsam, ``Geographic image retrieval using local invariant
  features,'' \emph{IEEE Transactions on Geoscience and Remote Sensing},
  vol.~51, no.~2, pp. 818--832, 2012.

\bibitem{risojevic2012fusion}
V.~Risojevi{\'c} and Z.~Babi{\'c}, ``Fusion of global and local descriptors for
  remote sensing image classification,'' \emph{IEEE Geoscience and Remote
  Sensing Letters}, vol.~10, no.~4, pp. 836--840, 2012.

\bibitem{cheng2015auto}
G.~Cheng, P.~Zhou, J.~Han, L.~Guo, and J.~Han, ``Auto-encoder-based shared
  mid-level visual dictionary learning for scene classification using very high
  resolution remote sensing images,'' \emph{IET Computer Vision}, vol.~9,
  no.~5, pp. 639--647, 2015.

\bibitem{zou2016scene}
J.~Zou, W.~Li, C.~Chen, and Q.~Du, ``Scene classification using local and
  global features with collaborative representation fusion,'' \emph{Information
  Sciences}, vol. 348, pp. 209--226, 2016.

\bibitem{zoph2016neural}
B.~Zoph and Q.~V. Le, ``Neural architecture search with reinforcement
  learning,'' \emph{arXiv preprint arXiv:1611.01578}, 2016.

\bibitem{zoph2018learning}
B.~Zoph, V.~Vasudevan, J.~Shlens, and Q.~V. Le, ``Learning transferable
  architectures for scalable image recognition,'' in \emph{Proceedings of the
  IEEE conference on computer vision and pattern recognition}, 2018, pp.
  8697--8710.

\bibitem{real2017large}
E.~Real, S.~Moore, A.~Selle, S.~Saxena, Y.~L. Suematsu, J.~Tan, Q.~V. Le, and
  A.~Kurakin, ``Large-scale evolution of image classifiers,'' in
  \emph{Proceedings of the 34th International Conference on Machine
  Learning-Volume 70}.\hskip 1em plus 0.5em minus 0.4em\relax JMLR. org, 2017,
  pp. 2902--2911.

\bibitem{real2019regularized}
E.~Real, A.~Aggarwal, Y.~Huang, and Q.~V. Le, ``Regularized evolution for image
  classifier architecture search,'' in \emph{Proceedings of the AAAI Conference
  on Artificial Intelligence}, vol.~33, 2019, pp. 4780--4789.

\bibitem{liu2018darts}
H.~Liu, K.~Simonyan, and Y.~Yang, ``Darts: Differentiable architecture
  search,'' \emph{arXiv preprint arXiv:1806.09055}, 2018.

\bibitem{xia2017aid}
G.-S. Xia, J.~Hu, F.~Hu, B.~Shi, X.~Bai, Y.~Zhong, L.~Zhang, and X.~Lu, ``Aid:
  A benchmark data set for performance evaluation of aerial scene
  classification,'' \emph{IEEE Transactions on Geoscience and Remote Sensing},
  vol.~55, no.~7, pp. 3965--3981, 2017.

\bibitem{chen2017deeplab}
L.-C. Chen, G.~Papandreou, I.~Kokkinos, K.~Murphy, and A.~L. Yuille, ``Deeplab:
  Semantic image segmentation with deep convolutional nets, atrous convolution,
  and fully connected crfs,'' \emph{IEEE transactions on pattern analysis and
  machine intelligence}, vol.~40, no.~4, pp. 834--848, 2017.

\bibitem{yu2015multi}
F.~Yu and V.~Koltun, ``Multi-scale context aggregation by dilated
  convolutions,'' \emph{arXiv preprint arXiv:1511.07122}, 2015.

\bibitem{szegedy2015going}
C.~Szegedy, W.~Liu, Y.~Jia, P.~Sermanet, S.~Reed, D.~Anguelov, D.~Erhan,
  V.~Vanhoucke, and A.~Rabinovich, ``Going deeper with convolutions,'' in
  \emph{Proceedings of the IEEE conference on computer vision and pattern
  recognition}, 2015, pp. 1--9.

\bibitem{he2016deep}
K.~He, X.~Zhang, S.~Ren, and J.~Sun, ``Deep residual learning for image
  recognition,'' in \emph{Proceedings of the IEEE conference on computer vision
  and pattern recognition}, 2016, pp. 770--778.

\bibitem{szegedy2016rethinking}
C.~Szegedy, V.~Vanhoucke, S.~Ioffe, J.~Shlens, and Z.~Wojna, ``Rethinking the
  inception architecture for computer vision,'' in \emph{Proceedings of the
  IEEE conference on computer vision and pattern recognition}, 2016, pp.
  2818--2826.

\bibitem{huang2017densely}
G.~Huang, Z.~Liu, L.~Van Der~Maaten, and K.~Q. Weinberger, ``Densely connected
  convolutional networks,'' in \emph{Proceedings of the IEEE conference on
  computer vision and pattern recognition}, 2017, pp. 4700--4708.

\bibitem{zagoruyko2016wide}
S.~Zagoruyko and N.~Komodakis, ``Wide residual networks,'' \emph{arXiv preprint
  arXiv:1605.07146}, 2016.

\bibitem{sandler2018mobilenetv2}
M.~Sandler, A.~Howard, M.~Zhu, A.~Zhmoginov, and L.-C. Chen, ``Mobilenetv2:
  Inverted residuals and linear bottlenecks,'' in \emph{Proceedings of the IEEE
  Conference on Computer Vision and Pattern Recognition}, 2018, pp. 4510--4520.

\end{thebibliography}

%\begin{thebibliography}{1}

%H.~Kopka and P.~W. Daly, \emph{A Guide to \LaTeX}, %3rd~ed.\hskip 1em plus

%\end{thebibliography}


%\end{IEEEbiography}


%\newpage



%\vfill


\end{document}